# DoF-Query Transformer for Adaptive Fusion of Egocentric Vision and Forearm HD-sEMG in Continuous Hand Kinematics

Jiahang Liu[1], Wenbo Zhang[1], Junsong Wang[1], Zhenzhi Ying[2*], Siqi Hou[1], Naohiko Sugita[2], Liming Shu[1*]

[1]School of Mechanical Engineering, Dalian University of Technology, Dalian, Liaoning, China.

[2]Department of Mechanical Engineering, The University of Tokyo, Tokyo, Japan.

**Abstract**

Continuous and accurate estimation of human hand kinematics is important not only for intuitive human–machine interaction, but also for capturing fine-grained human manipulation demonstrations that can provide action supervision for embodied-intelligence datasets and dexterous robot learning. However, individual sensing modalities may be insufficient for reliable estimation under varying operating conditions. We present a finger-unencumbered multimodal framework for continuous regression of 20 hand joint angles from egocentric RGB images and 128-channel forearm high-density surface electromyography (HD-sEMG). The framework combines a Vision Transformer-based visual encoder, a Transformer-based EMG encoder, and a degree of freedom (DoF) query Transformer fusion decoder. To account for the varying informativeness of visual and EMG cues across joint DoFs and input conditions, the decoder assigns a separate learnable query to each DoF, enabling DoF-specific and context-adaptive fusion of the two modalities. The framework was evaluated on synchronized multimodal data from 10 participants. Averaged across participants, the proposed fusion model achieved a mean coefficient of determination ($R^2$) of 0.88 and a mean absolute error (MAE) of 2.80°, compared with 0.82 and 3.21° for the vision-only model and 0.68 and 5.13° for the EMG-only model. It also outperformed the evaluated representative unimodal baselines and alternative fusion designs. DoF-wise analyses showed larger fusion gains for visually ambiguous joints and viewing conditions, while synthetic visual occlusion demonstrated greater robustness than vision-only estimation. These results highlight the complementary value of egocentric vision and forearm HD-sEMG for capturing fine-grained human hand kinematics, with potential applications in human demonstration acquisition and dexterous robot learning.

## 1. INTRODUCTION

Human hands are among the primary means by which people manipulate objects, convey intentions, and interact with physical and digital environments. Accurate and continuous estimation of hand kinematics can facilitate intuitive human–robot interaction (HRI), teleoperation, dexterous robotic manipulation, and augmented and virtual reality (AR/VR) applications. Beyond online interaction and control, fine-grained hand-kinematic trajectories also provide a compact representation of how humans execute manipulation behaviors. When synchronized with egocentric visual observations, these trajectories can serve as dense action annotations for human demonstration data, providing valuable supervision for embodied-intelligence datasets and data-driven models of dexterous manipulation. After appropriate kinematic retargeting, such human hand-motion data may further support imitation learning and policy learning for robotic hands. Many of these applications require continuous estimates of hand joint angles rather than classification into a predefined set of gesture categories. Compared with discrete gesture recognition, joint-angle regression preserves fine-grained kinematic information and provides a more direct mapping from human hand movements to robotic or interactive systems.

Hand-worn sensors have been widely used to capture hand kinematics. Their close physical coupling with the hand can provide informative and direct measurements. However, devices mounted on the fingers or palm may interfere with natural hand movements, reduce user comfort, and constrain everyday interactions. Piezoresistive bend sensors may exhibit time-dependent signal drift, hysteresis, and sensitivity degradation after repeated deformation [1]. Capacitive sensors generally offer low mechanical hysteresis, but they can be affected by parasitic capacitance, environmental conditions, and electromagnetic interference, potentially limiting the accuracy of subtle finger-motion measurements [2]. Optical-fiber sensors may require dedicated optical interrogation hardware, and some designs rely on reference or calibration channels to compensate for environmental variations and maintain measurement accuracy [3]. Inertial measurement units (IMU)-based gloves are susceptible to orientation drift caused by sensor noise and integration errors, while magnetometer-assisted systems may also be affected by magnetic disturbances [4]. Hand exoskeletons are often rigid, may require user-specific fitting or calibration, and can measure only the degree of freedom (DoF) permitted by their mechanical design, making them less suitable for natural and unconstrained interaction [5]. These limitations are particularly relevant when hand kinematics are used to collect human manipulation demonstrations, because sensing devices attached to the fingers or palm can alter natural contact, grasping, and object-manipulation behaviors. A finger-unencumbered sensing approach is therefore desirable not only for user comfort, but also for acquiring hand-motion data that more faithfully reflect natural manipulation.

Vision-based hand pose estimation has advanced substantially in recent years. Marker-based optical motion capture can provide highly accurate 3D kinematic reference measurements, while markerless vision systems (e.g. DexPilot, FreiHAND, and HandDiff) enable hand tracking without requiring sensors to be attached to the hand [6-8]. However, multi-view and depth-assisted systems may require calibrated camera arrangements or constrained capture environments, which can limit their applicability in mobile and everyday interaction scenarios. Monocular RGB cameras provide rich appearance and 2D spatial cues while remaining compact and easy to deploy. Egocentric monocular configurations are therefore particularly attractive for wearable applications. Recent methods have considerably improved 3D hand reconstruction from single RGB images. HaMeR uses a Vision Transformer-Huge (ViT-H) backbone and a cross-attention Transformer decoder for 3D hand reconstruction [9]. Hamba uses graph-guided state-space blocks with bidirectional Mamba scanning of joint-sampled tokens to model joint relations for single-view 3D hand reconstruction [10]. WiLoR predicts an initial hand pose from image and learnable tokens, then refines it using image-aligned multiscale features [11]. Nevertheless, monocular RGB-based estimation remains an inherently ill-posed problem and is susceptible to self-occlusion, finger overlap, foreshortening, motion blur, and depth ambiguity. These challenges can be particularly pronounced in egocentric settings. Consequently, RGB-only methods may produce inaccurate or temporally unstable joint-angle estimates under challenging viewpoints, severe occlusion, or rapid hand motion.

Surface electromyography (sEMG) measures the electrical activity of muscles associated with finger movements without requiring the hand to remain visible. It can therefore provide information about muscle activation associated with ongoing or intended movements even when the hand is occluded. Previous studies have investigated wearable sEMG for hand-kinematics estimation. NeuroPose demonstrated the feasibility of tracking 3D finger motion using wearable EMG sensors [12]. The emg2pose benchmark provides a large-scale dataset of synchronized wrist-based sEMG recordings and hand-pose labels obtained using a motion-capture system [13]. In addition, high-density sEMG (HD-sEMG) systems have achieved promising performance in complex hand-gesture classification and continuous joint-angle estimation [14]. Because many extrinsic muscles controlling the fingers are located in the forearm, sEMG electrodes can be placed around the wrist or forearm without obstructing the fingers, making this modality attractive for wearable and relatively unconstrained interaction. Nevertheless, sEMG does not provide explicit information about hand geometry or absolute joint configuration. Neither monocular RGB vision nor sEMG alone is therefore sufficient to ensure accurate and robust continuous hand joint-angle estimation under all operating conditions.

To mitigate the limitations of individual sensing modalities, recent studies have investigated multimodal fusion for hand pose and kinematic estimation. These approaches integrate complementary measurements from sensors such as data gloves, cameras, and IMUs to improve robustness to modality-specific failure modes, including visual occlusion, inertial drift, and environmental disturbances [15, 16]. RGB images provide information about hand appearance and observable spatial configuration, whereas sEMG captures forearm muscle activation and can remain informative when visual observations are degraded by occlusion or motion blur. Existing studies on EMG–vision fusion have largely focused on discrete gesture classification [17, 18]. Consequently, the fusion of egocentric RGB images and HD-sEMG signals for fine-grained, continuous hand joint-angle regression remains underexplored. Moreover, the relative informativeness of the two modalities may vary across degrees of freedom (DoFs), hand configurations, visibility conditions, and phases of motion. A global and time-invariant fusion strategy may therefore be suboptimal.

Based on these observations, we propose a finger-unencumbered multimodal sensing framework that combines egocentric RGB images with 128-channel HD-sEMG signals recorded from the forearm to continuously estimate 20-DoF hand joint angles. We employ a Vision Transformer (ViT)-based visual encoder to extract hand appearance and spatial features from egocentric images and a Transformer-based EMG encoder to model the spatiotemporal dependencies within the HD-sEMG signals. To integrate the two modalities, we further introduce a DoF-query Transformer fusion decoder. The decoder assigns an independent learnable query to each hand DoF. At each time step, each query adaptively attends to relevant visual and EMG tokens to estimate its corresponding joint angle. We evaluate the proposed framework on a synchronized multimodal dataset. The proposed fusion model consistently outperforms the vision-only and EMG-only models, alternative multimodal fusion baselines, and representative state-of-the-art methods. Under simulated visual occlusion, the DoF-query Transformer exhibits more graceful performance degradation than the other fusion methods and the vision-only model, demonstrating improved robustness to degraded visual input. The proposed framework provides a sensing pathway for obtaining dense, continuous hand-kinematic annotations from finger-unencumbered human demonstrations.

The main contributions of this work are summarized as follows:

1. We develop and evaluate a finger-unencumbered multimodal sensing framework that combines egocentric RGB images with 128-channel forearm HD-sEMG signals for continuous 20-DoF hand joint-angle regression, providing fine-grained kinematic representations of human hand movements without instrumenting the fingers.

2. We propose a DoF-query Transformer fusion decoder that integrates visual and EMG representations through independent per-DoF queries, enabling DoF-specific and context-adaptive fusion for continuous hand joint-angle estimation.

3. Through experiments on a synchronized multimodal dataset, we demonstrate that the proposed framework outperforms single-modality models and alternative fusion baselines, while maintaining greater robustness under simulated visual occlusion.

## 2. METHODS

This study proposes a finger-unencumbered multimodal DoF-query fusion framework that jointly leverages forearm HD-sEMG and egocentric RGB images to continuously estimate a 20-DoF hand joint-angle representation. As shown in Fig. 1, the framework consists of two parallel modality-specific branches, each comprising a preprocessing module and an encoder network. The outputs of the two branches are then fed into a DoF-query fusion decoder followed by a shared regression head. The decoder assigns an independent query to each hand DoF and aggregates the visual and EMG representations in a DoF-specific and context-adaptive manner, so that each joint angle can selectively integrate visual and EMG information according to the current context.

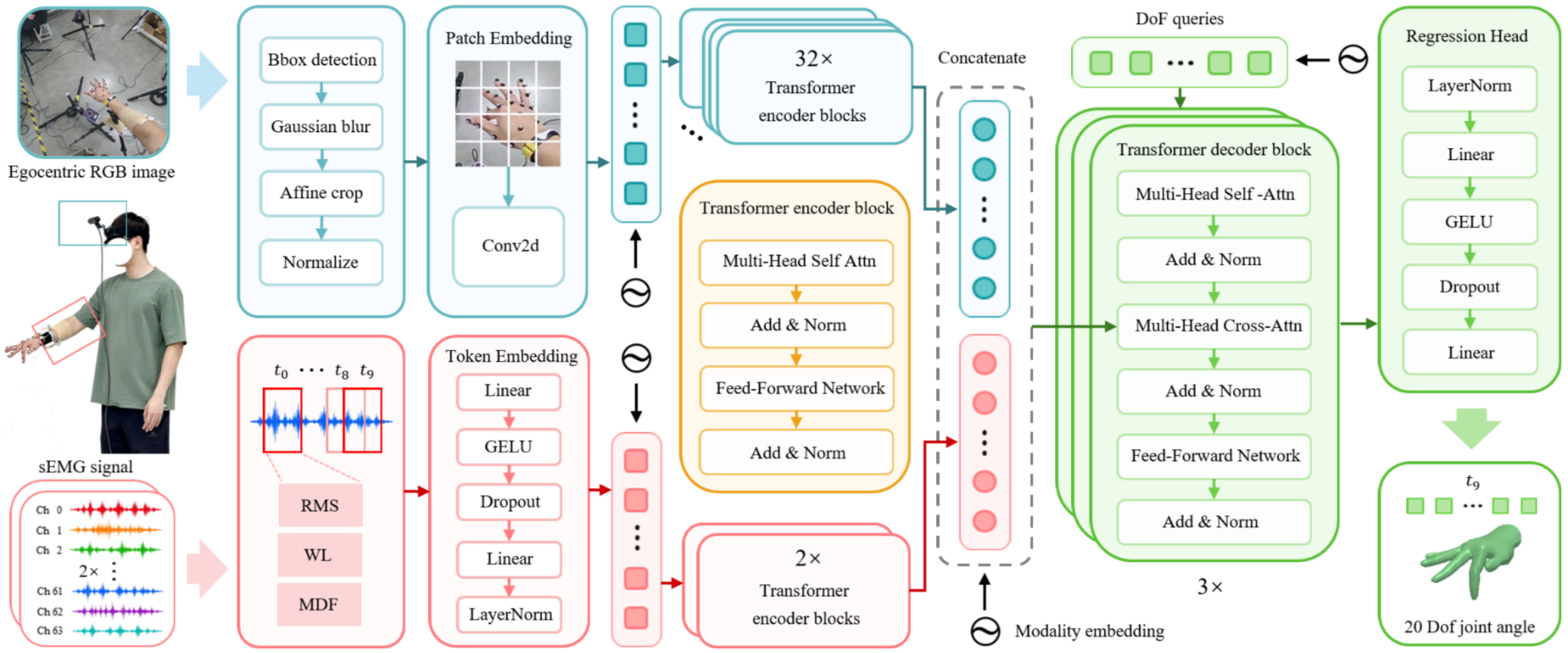

*__Fig. 1__. Overview of the proposed multimodal fusion framework. The framework takes synchronized egocentric RGB images and forearm HD-sEMG signals as inputs. The vision branch extracts spatial hand-region tokens from cropped RGB patches using a ViT backbone, while the EMG branch converts windowed hand-crafted sEMG features into temporal EMG tokens using a Transformer encoder. After adding modality embeddings, tokens from both modalities are concatenated and fused by a query-based Transformer decoder, in which each hand DoF is represented by an independent learnable query and mapped to its corresponding joint angle through a shared multilayer perceptron (MLP) regression head for continuous estimation of 20 hand joint angles.*

## 2.1 Vision Branch

As illustrated in Fig. 1, the visual branch, highlighted in blue, extracts visual representations of the hand region from the RGB images. We select the RGB image frame corresponding to the end time step of each EMG feature sequence as the visual input. To reduce the influence of background regions on hand feature extraction, we first apply ViTDet [19] to detect the hand region in the image and obtain the corresponding hand bounding box. Based on the detected bounding box, the hand region in the original image is then cropped and preprocessed. We first expand the bounding box with a rescale factor of 2.0 to preserve sufficient hand context and avoid an overly tight crop. The expanded hand region is then affine-transformed into a square image patch with a resolution of 256×256. When the down-sampling ratio is large, Gaussian smoothing is applied before generating the image patch to reduce aliasing artifacts. The patch is subsequently normalized using the ImageNet mean and standard deviation. The resulting visual input can be denoted as

$$X_{\text{vis}} \in \mathbb{R}^{B\times 3\times H\times W} \tag{1}$$

where $B$ is the batch size, and $H = W = 256$ denote the height and width of the cropped hand image patch.

In this study, we adopt the ViT-H backbone from HaMeR as the visual feature extractor. Through pretraining on large-scale hand data, this model exhibits strong robustness and generalization in monocular 3D hand mesh recovery, so its parameters are initialized with the HaMeR pretrained weights. To match the input resolution used by the pretrained HaMeR ViT-H backbone, the preprocessed 256×256 hand patch is center-cropped along the width dimension by removing 32 pixels from both the left and right sides. This yields a 256×192 image that is subsequently fed into the visual backbone. The image is then partitioned by a 2D convolutional neural network (CNN) patch embedding layer (kernel size = 16×16, stride = 16×16, padding = 2), projected into a 1280-dimensional embedding space, and flattened into 16×12 = 192 visual tokens:

$$T_{\text{vis}} \in \mathbb{R}^{B\times N\times d_{\text{model}}} \tag{2}$$

where $B$ is the batch size, $N = 192$ denotes the number of visual tokens, and $d_{\text{model}} = 1280$ denotes the hidden dimension of each token.

To preserve spatial position information, we add learnable absolute positional embeddings to the visual token sequence. The visual token sequence is then fed into a Transformer encoder to model the spatial dependencies among different patches. The visual encoder consists of 32 Transformer encoder blocks, each with 16 attention heads, a hidden dimension of 1280, and a feed-forward network dimension of 5120. Stochastic depth is applied across the encoder blocks, with the drop path rate increasing linearly from 0 to 0.55.

## 2.2 EMG Branch

The EMG branch, shown in red in Fig. 1, processes 128-channel HD-sEMG signals acquired from the forearm using two 8 × 8 electrode arrays. The raw HD-sEMG signals were first filtered to reduce motion artifacts, high-frequency noise, and power-line interference. Specifically, a 10–450 Hz band-pass filter was applied to suppress low- and high frequency noise, while notch filters at 50 Hz and its harmonics within the passband were used to suppress power-line interference. The filtered signals were then segmented into consecutive windows using a sliding-window strategy. A window length of 200 ms and a step size of 100 ms were used. The resulting windowed EMG input can be denoted as

$$X_{\mathrm{EMG}} \in \mathbb{R}^{B\times S\times W\times C} \tag{3}$$

where $B$ is the batch size and $S$ denotes the number of consecutive EMG windows in the input sequence. In this study, 10 adjacent EMG windows were used as one sequence to characterize the temporal variation of the EMG signals. W is the number of samples within each window. Since the EMG sampling rate in this study is 2000 Hz, a 200 ms window corresponds to W = 400 samples. C denotes the 128 EMG channels.

The windowed EMG signals were then used for hand-crafted feature extraction. For each temporal window and each EMG channel, three types of hand-crafted features were extracted: root mean square (RMS), waveform length (WL), and median frequency (MDF). Therefore, after hand-crafted feature extraction, each EMG channel within each temporal window was represented by a feature vector with a length of 3. The resulting EMG feature tensor can be denoted as

$$F_{\mathrm{EMG}} \in \mathbb{R}^{B\times S\times 3\times C} \tag{4}$$

Before being fed into the EMG encoder, the hand-crafted features were normalized using z-score normalization to reduce the influence of scale differences across feature types and channels on model training. Normalization was performed independently along the temporal dimension for each feature–channel combination. All normalization statistics were computed only from the training set and then fixed during validation and testing, thereby avoiding information leakage from validation or test data.

The normalized EMG feature tensor was then fed into the Transformer-based EMG encoder. Each temporal window contained three types of hand-crafted features across 128 EMG channels, which were concatenated into an overall feature vector for each temporal window. This feature vector was then mapped to a unified hidden dimension through a nonlinear token embedding module. The embedding module consisted of a linear layer, GELU activation, dropout, a second linear layer, and LayerNorm, thereby transforming the original hand-crafted feature vector into a high-dimensional token representation suitable for Transformer processing. The embedded EMG token sequence can be denoted as

$$T_{\mathrm{EMG}} \in \mathbb{R}^{B\times S\times d_{\mathrm{model}}} \tag{5}$$

where $d_{\mathrm{model}}$ was set to 1280. To preserve the temporal position information of different EMG windows in the sequence, a learnable one-dimensional temporal positional embedding was further added. The EMG token sequence was then fed into a Transformer encoder to model the dynamic changes between consecutive EMG windows along the temporal dimension. The encoder consisted of two Transformer encoder blocks. Each block consisted of a multi-head self-attention module and a feed-forward network, with residual connections, LayerNorm, and dropout used for stable training. Each layer used 8 attention heads, the hidden dimension was 1280, the feed-forward network dimension was 2560, and the dropout rate was set to 0.1.

### 2.3 DoF-query Transformer Fusion Decoder

For hand joint angle prediction, the reliability and relevance of EMG and visual features are not globally consistent, but vary across DoFs, hand configurations, visibility conditions, and phases of motion. Directly concatenating or pooling the two modalities into a single global representation may therefore mix irrelevant or unreliable information and weaken joint and context specific discriminative cues. To address this issue, we employ a query-based Transformer decoder inspired by DETR [20], as highlighted in green in Fig. 1. In DETR, learnable object queries act as detection slots that attend to encoded image features to produce object-level representations. In our formulation, a separate learnable query is assigned to each hand DoF. These DoF queries attend to the multimodal memory to obtain DoF-specific representations for joint-angle prediction.

Before constructing the decoder memory, learnable modality embeddings are added to the corresponding modality tokens, and the visual and EMG token sequences are concatenated along the token dimension:

$$M = \text{Concatenate}(T_{\text{EMG}} + e_{\text{EMG}}, T_{\text{vis}} + e_{\text{vis}}) \in \mathbb{R}^{B\times(S+N)\times d_{\text{model}}} \quad (6)$$

Here, $M$ denotes the decoder memory. $T_{EMG}$ and $T_{vis}$ denote the EMG and visual tokens, respectively. $e_{EMG}$ and $e_{vis}$ are learnable modality embeddings for the EMG and visual modalities, respectively. In this way, each token in the memory carries both its feature representation and modality identity. For $J = 20$ hand DoFs, we define a set of learnable DoF-query tokens $\mathbf{Q}^0 \in \mathbb{R}^{J\times d_{model}}$, which are then fed into a stack of Transformer decoder layers.

Each Transformer decoder layer contains three sequential submodules: multi-head self-attention, multi-head cross-attention, and a feed-forward network (FFN). Given the input matrices $\mathbf{X}_Q$, $\mathbf{X}_K$, and $\mathbf{X}_V$, the multi-head attention operation is defined as

$$\text{MultiHeadAttention}(\mathbf{X}_Q, \mathbf{X}_K, \mathbf{X}_V) = \text{Concat}(\text{head}_1, \dots, \text{head}_H)\mathbf{W}^O \quad (7)$$

where

$$\text{head}_h = \text{Attn}(\mathbf{X}_Q\mathbf{W}_h^Q, \mathbf{X}_K\mathbf{W}_h^K, \mathbf{X}_V\mathbf{W}_h^V) \quad (8)$$

Here, $\mathbf{W}_h^Q$, $\mathbf{W}_h^K$, and $\mathbf{W}_h^V$ are the query, key, and value projection matrices of the $h$-th attention head. The outputs from all attention heads are concatenated along the feature dimension and projected by $\mathbf{W}^O$. The single-head attention operation is defined as

$$\text{Attn}(\mathbf{Q}_h, \mathbf{K}_h, \mathbf{V}_h) = \mathbf{A}_h\mathbf{V}_h = \text{Softmax}\left(\frac{\mathbf{Q}_h\mathbf{K}_h^T}{\sqrt{d_h}}\right)\mathbf{V}_h \quad (9)$$

where $\mathbf{Q}_h = \mathbf{X}_Q\mathbf{W}_h^Q$, $\mathbf{K}_h = \mathbf{X}_K\mathbf{W}_h^K$, and $\mathbf{V}_h = \mathbf{X}_V\mathbf{W}_h^V$ are the projected query, key, and value representations, respectively. The matrix $\mathbf{A}_h$ denotes the normalized attention-weight matrix. In our implementation, the number of attention heads is $H = 8$, the model dimension is $d_{\text{model}} = 1280$, and the feature dimension of each head is $d_h = d_{\text{model}}/H = 160$. For the $l$-th layer in a stack of $L = 3$ Transformer decoder layers, the post-norm update is formulated as:

$$\hat{\mathbf{Q}}^{(l)} = \text{LayerNorm}\left(\mathbf{Q}^{(l-1)} + \text{MultiHeadAttention}_{\text{self}}\left(\mathbf{Q}^{(l-1)}, \mathbf{Q}^{(l-1)}, \mathbf{Q}^{(l-1)}\right)\right) \tag{10}$$

$$\overline{\mathbf{Q}}^{(l)} = \text{LayerNorm}\left(\hat{\mathbf{Q}}^{(l)} + \text{MultiHeadAttention}_{\text{cross}}\left(\hat{\mathbf{Q}}^{(l)}, \mathbf{M}, \mathbf{M}\right)\right) \tag{11}$$

$$\mathbf{Q}^{(l)} = \text{LayerNorm}\left(\overline{\mathbf{Q}}^{(l)} + \text{FFN}\left(\overline{\mathbf{Q}}^{(l)}\right)\right) \tag{12}$$

Here, $\mathbf{Q}^{(l-1)}$ denotes the query representation input to the $l$-th decoder layer. The self-attention module operates on this representation and models dependencies among different DoF queries, producing the intermediate representation $\hat{\mathbf{Q}}^{(l)}$. The cross-attention module then uses $\hat{\mathbf{Q}}^{(l)}$ as the query input and enables each DoF-query to selectively aggregate relevant information from the multimodal memory $M$, resulting in the conditioned representation $\overline{\mathbf{Q}}^{(l)}$. Finally, the FFN independently transforms each DoF-specific representation and produces the output $\mathbf{Q}^{(l)}$ of the current decoder layer. The hidden layer dimension of FFN is 2560.

For interpretability analysis, we use the cross-attention weights from the final decoder layer, indexed by $l = 3$. The first $S$ memory tokens correspond to EMG features, whereas the remaining $N$ tokens correspond to visual features. For the $j$-th DoF of the $i$-th sample, the modality-specific attention allocations are computed as

$$m_{i,j}^{\text{EMG}} = \frac{1}{H} \sum_{h=1}^{H} \sum_{s=1}^{S} A_{i,j,s}^{(3,h)} \tag{13}$$

$$m_{i,j}^{\text{vis}} = \frac{1}{H} \sum_{h=1}^{H} \sum_{n=1}^{N} A_{i,j,S+n}^{(3,h)} \tag{14}$$

Here, the element $A_{i,j,r}^{(l,h)}$ measures the attention assigned by the $j$-th DoF-query of the $i$-th sample to the $r$-th memory token. Specifically, $l = 3$ indexes the decoder layer and $h$ indexes the attention head. Because the attention weights are normalized over all memory tokens, these two quantities satisfy

$$m_{i,j}^{\text{EMG}} + m_{i,j}^{\text{vis}} = 1 \tag{15}$$

These quantities characterize the DoF-specific and input-dependent allocation of cross-attention mass between the two modalities. A larger value indicates that a greater fraction of the cross-attention mass is assigned to tokens from the corresponding modality under the current input condition.

A shared regression head is applied independently to each DoF representation. For the $j$-th DoF, the predicted joint angle is given by

$$\hat{y}_j = \mathbf{W}_2 \, \text{Dropout}\left[\text{GELU}\left(\mathbf{W}_1 \, \text{LayerNorm}\left(\mathbf{z}_j\right) + \mathbf{b}_1\right)\right] + b_2 \tag{16}$$

Here, $\mathbf{z}_j \in \mathbb{R}^{d_{\text{model}}}$ denotes the final representation associated with the $j$-th DoF. The LayerNorm operation normalizes the feature dimensions of each DoF representation independently. $\mathbf{W}_1$ and $\mathbf{b}_1$ parameterize the first fully connected layer. $\mathbf{W}_2$ and $\mathbf{b}_2$ parameterize the final linear projection from the hidden representation to a scalar output. GELU is used as the nonlinear activation function, and dropout with a probability of 0.1 is applied during training. The same regression-head parameters are shared across all DoFs, while the head is applied independently to each DoF-specific representation.

## 2.4 Experimental Setup

In this study, we collected synchronized multimodal data, including egocentric RGB images, HD-sEMG signals, and three-dimensional trajectories of hand joint motion-capture markers, for training and evaluating the proposed fusion model. We used the 3D marker trajectories obtained from the optical motion capture system to compute the ground-truth hand joint angles during hand movements. Ten healthy right-handed adults participated in this experiment. All participants provided written informed consent prior to participation. The experimental protocol was conducted in accordance with the Declaration of Helsinki and was approved by the Ethics Committee of Dalian University of Technology.

As shown in Fig. 2(A), each participant stood within the experimental area during data acquisition, with the right wrist resting on a tripod support to reduce arm fatigue during the prolonged experiment and to help maintain a stable forearm position. An RGB-D camera (Orbbec Gemini 335L), with only its RGB stream enabled, was mounted on the participant's head to capture egocentric images of hand movements throughout the experiment. The camera was positioned to ensure that the participant's hand movements were within the field of view. RGB images were recorded at a resolution of 1280 × 720 pixels and a frame rate of 30 Hz.

HD-sEMG signals were recorded from the participant's right forearm using a self-developed wireless HD-sEMG acquisition system. As shown in Fig. 2(B), each electrode-array channel was differentially measured with respect to a reference electrode mounted at the wrist. The acquisition unit comprised power-management, EMG signal-processing, and wireless-communication modules, including a microprocessor for control of signal modulation and demodulation. The differential signals between the electrode-array channels and the common reference electrode were successively amplified, band-pass filtered and digitized. The processed signals were then transmitted to the host computer through a radio-frequency WIFI interface. Charging contacts were provided for battery charging. As shown in Fig. 2(C), two wireless acquisition units from the HD-sEMG acquisition system were used in the experiment. Each acquisition unit was connected to a 64-channel distributed 8 × 8 square electrode arrays, providing a total of 128 recording channels. The two electrode arrays were placed on the dorsal and volar sides of the right forearm to record surface EMG signals from the extensor and flexor muscle groups, respectively. The reference electrode was mounted at the wrist and served as the common reference for differential recording by the two acquisition units. Conductive gel was applied at the electrode–skin interface to reduce contact impedance and improve signal quality. The electrode arrays were first attached to the forearm skin using their adhesive backing and then further secured with bandages to maintain stable electrode–skin contact and minimize electrode displacement and motion artifacts during data acquisition. The HD-sEMG signals were sampled at 2000 Hz.

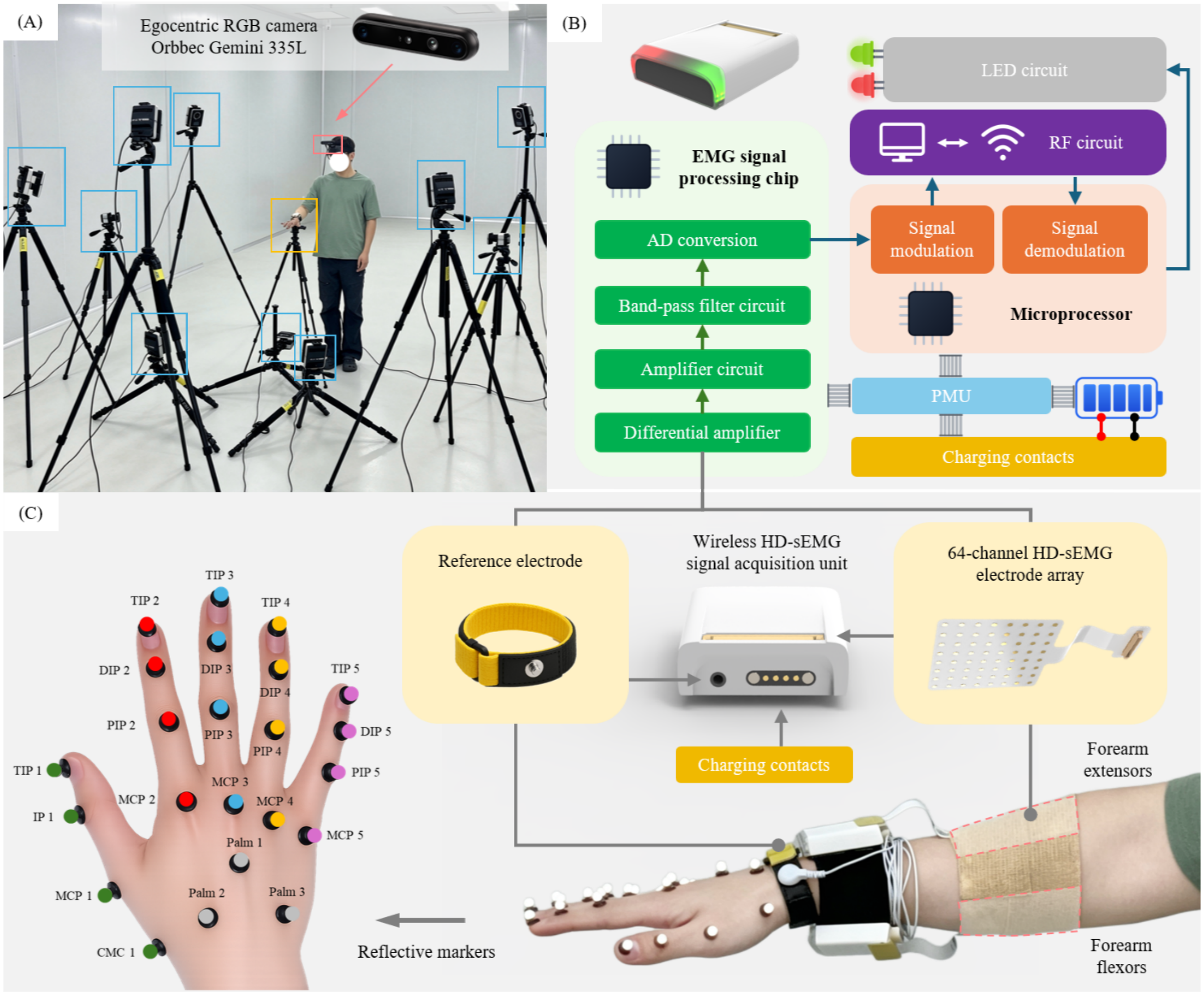

***Fig. 2****. Experimental setup for multimodal data acquisition. (A) An RGB camera was set on the forehead of the participant. 10 motion-capture cameras were arranged around the experimental area to record the hand movements of the participant. The red, blue, and yellow boxes indicate the egocentric RGB camera, motion-capture cameras, and the hand–forearm sensing setup, respectively. (B) Functional schematic of the wireless HD-sEMG acquisition unit. The unit comprises three main functional parts. The power-management unit (PMU) supplies power to the acquisition and processing circuits. The EMG signal-processing path receives the differential signals generated between the electrode-array channels and the wrist-mounted common reference electrode, and performs amplification, band-pass filtering, analog-to-digital conversion, and on-board signal processing. The microprocessor controls signal modulation and demodulation and communicates with the host computer through the radio-frequency wireless interface. (C) Placement of motion-capture markers and the HD-sEMG acquisition system. Reflective markers were attached to anatomical landmarks on the dorsal hand to record hand kinematics, with three palm markers defining the palm reference frame. Two 64-channel HD-sEMG electrode arrays were placed over the forearm flexor and extensor regions, respectively. Each array was connected to a wireless acquisition unit, and a wrist-mounted reference electrode provided the common reference for differential recording.*

To obtain ground-truth hand joint angles, an optical motion capture system with 10 cameras (Vicon Motion Systems Ltd) was used to record the three-dimensional trajectories of reflective markers attached to the right hand [21]. As shown in Fig. 2(A), the cameras were arranged around the participant to provide coverage of the hand movement workspace and improve marker visibility across different hand gestures and postures. Reflective markers were attached to anatomical landmarks on the dorsal surface of the right hand, as shown in Fig. 2(C). The fingers were numbered from 1 to 5, corresponding to the thumb, index, middle, ring, and pinky fingers, respectively. Three markers, named Palm 1, Palm 2, and Palm 3 in counterclockwise order, were attached to the dorsal surface of the hand to define the palm reference frame. For the thumb, four markers were placed at anatomical landmarks corresponding to the carpometacarpal (CMC), metacarpophalangeal (MCP), interphalangeal (IP), and fingertip (TIP) locations. For the index, middle, ring, and pinky fingers, four markers were placed at anatomical landmarks corresponding to the metacarpophalangeal (MCP), proximal interphalangeal (PIP), distal interphalangeal (DIP), and fingertip (TIP) locations of each finger. The motion capture data were recorded at 100 Hz. The 20-DoF hand joint-angle sequences were computed from the three-dimensional marker trajectories and expressed in degrees. The detailed formulas used for joint angle calculation are provided in Section 1 of the Supplementary Material. For model training, the joint-angle values of each DoF were standardized using z-score normalization, with the mean and standard deviation computed exclusively from the corresponding training set.

Each participant performed the experimental tasks under two hand orientations: palm-facing-down and palm-facing-side. Under each orientation, the participant completed 10 trials of predefined-gesture protocol and 20 trials of unconstrained free hand movements. Each trial lasted approximately 90 s, with a 60 s rest interval between consecutive trials to reduce fatigue. The predefined gesture movements were designed to collect representative HD-sEMG signals under different hand movement synergies, thereby providing initial coverage of distinct muscle activation patterns. The detailed predefined gesture set is shown in Supplementary Fig. 1. During the unconstrained free-movement trials, participants were encouraged to perform continuous and diverse hand movements involving different joint DoFs and movement synergies, rather than repeatedly performing a single isolated gesture. Across both orientations, each participant therefore completed 20 predefined-gesture trials and 40 free-movement trials. The 40 trials of unconstrained free hand movements were divided according to hand orientation. Under each orientation, one free-movement trial was randomly selected for the validation set, nine free-movement trials were randomly selected for the test set, and the remaining ten free-movement trials, together with the 10 predefined gesture trials, were used as the training set. Thus, the validation and test sets consisted exclusively of held-out free-movement trials.

During training, we used the Huber loss with delta = 1.0 for joint-angle regression. To account for the different initialization schemes of the model components, separate parameter groups were used with component-specific learning rates. The ViT-Huge visual backbone, initialized with the pretrained HaMeR weights, was fine-tuned with a learning rate of $1 \times 10^{-6}$, whereas the EMG encoder, DoF-query fusion decoder, and regression head were trained with a learning rate of $1 \times 10^{-4}$. Within each parameter group, decoupled weight decay was applied selectively: one-dimensional parameters, bias terms, and normalization-layer parameters were excluded from weight decay, whereas a weight-decay coefficient of $5 \times 10^{-4}$ was applied to the remaining parameters. The AdamW optimizer was used. A ReduceLROnPlateau scheduler monitored the validation loss, and all models were trained for 50 epochs with a batch size of 256. Model training and validation were performed on an NVIDIA GeForce RTX 5080 GPU, while auxiliary computations were performed on an Intel Core Ultra 9 275HX CPU.

## 3. RESULTS AND DISCUSSION

To comprehensively evaluate the accuracy and robustness of the proposed multimodal hand joint-angle regression framework, we conducted experiments on multimodal data collected from ten participants, including synchronized 128-channel HD-sEMG signals, egocentric RGB images, and 20-DoF hand joint angle labels derived from optical motion capture. We first investigated the contribution of each sensing modality by comparing the proposed fusion model with vision-only and EMG-only variants in terms of overall regression accuracy, temporal consistency, qualitative hand mesh reconstruction, and DoF-wise performance under different palm orientations. We then compared the proposed method with representative EMG-based and vision-based unimodal baselines to validate its advantage over existing single-modality approaches. Next, module-level ablation studies were performed to examine the effects of different EMG encoder architectures and fusion decoder designs. Finally, DoF-specific query analyses and synthetic visual-occlusion experiments were conducted to characterize the model's modality reliance across hand joint DoFs and to evaluate its robustness to degraded visual input. Model performance was quantitatively evaluated using the coefficient of determination ($R^2$) and the mean absolute error (MAE). $R^2$ measures the proportion of variance in the observed values that can be explained by the model, with values closer to 1 indicating better goodness of fit. MAE represents the average absolute difference between observed and predicted values and provides an intuitive measure of the average prediction error.

### 3.1 Modality Ablation

To assess the contribution of each sensing modality and evaluate the effectiveness of multimodal fusion, we compared three model variants for each participant: an EMG-only model, a vision-only model, and the proposed multimodal fusion model. The variants were trained and tested using the same data splits. For the unimodal variants, the decoder architecture and training configuration were identical to those of the fusion model, whereas only the encoder associated with the available modality was retained. We evaluated overall regression accuracy, temporal consistency of the predicted 20-DoF joint-angle trajectories, qualitative hand-mesh reconstruction, and DoF-wise performance across different palm orientations.

Fig. 3 shows the overall comparison of models with different input modalities using the $R^2$ score and MAE. The fusion model achieved the highest $R^2$ score and the lowest MAE, with 0.88 / 2.80°, compared with 0.82 / 3.21° for the vision-only model and 0.68 / 5.13° for the EMG-only model. The fusion model also shows the smallest max–min variation range across participants, indicating more accurate and stable estimation of hand joint angles.

Fig. 4 shows the temporal trajectories predicted by models with different input modalities for 20 hand joint DoFs in a palm-facing-down test sequence. Each subplot corresponds to one hand joint DoF: thumb IP and MCP, thumb CMC abduction and flexion, and the DIP, PIP, MCP flexion, and MCP abduction DoFs of the index, middle, ring, and pinky fingers. In the displayed sequence, the fusion model generally showed closer agreement with the ground-truth trajectories than either single-modality model, with smaller amplitude deviations for most DoFs.

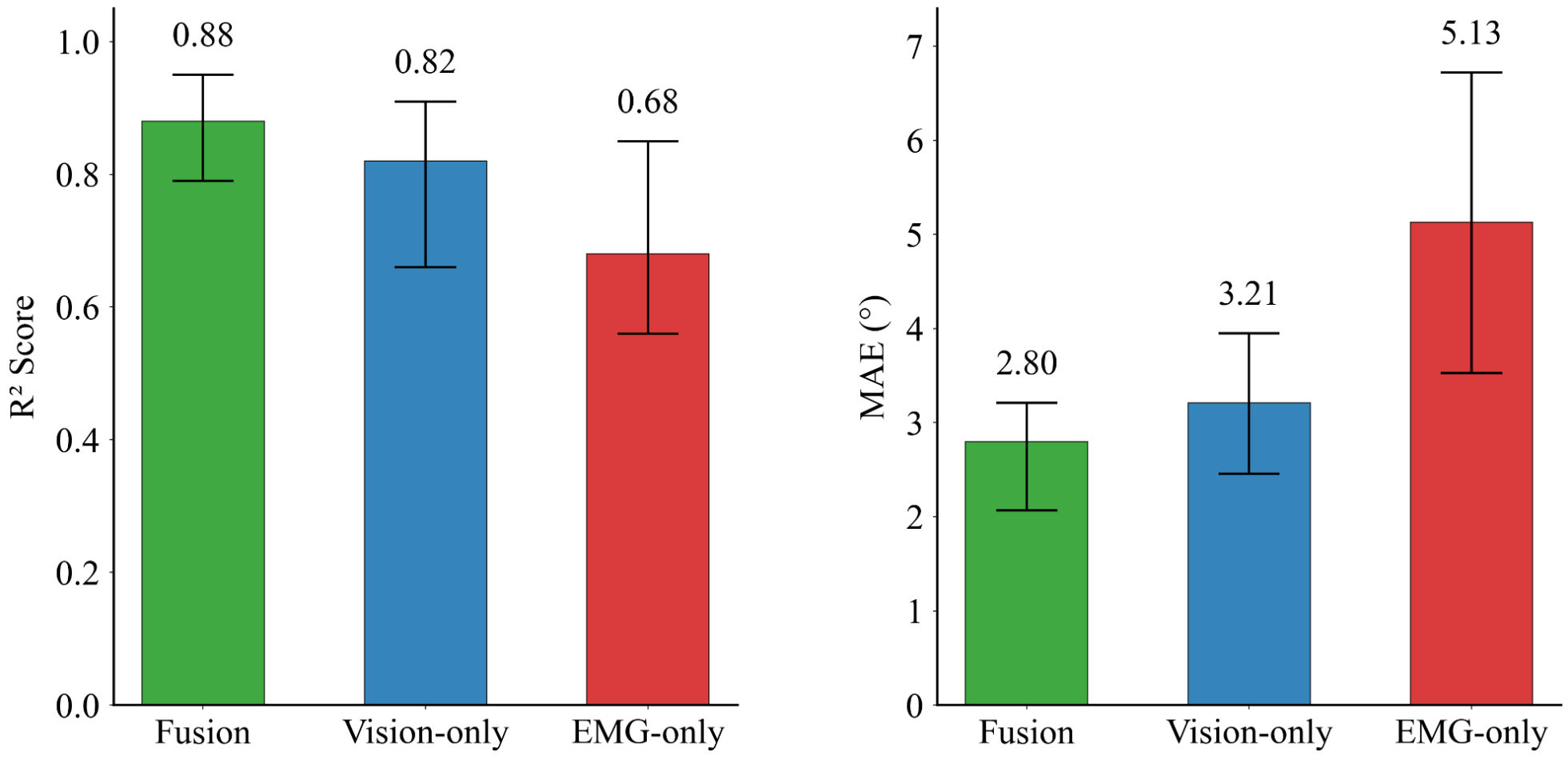

***Fig. 3***. *Overall comparison of fusion, vision-only, and EMG-only models using R² and MAE (°) metrics.*

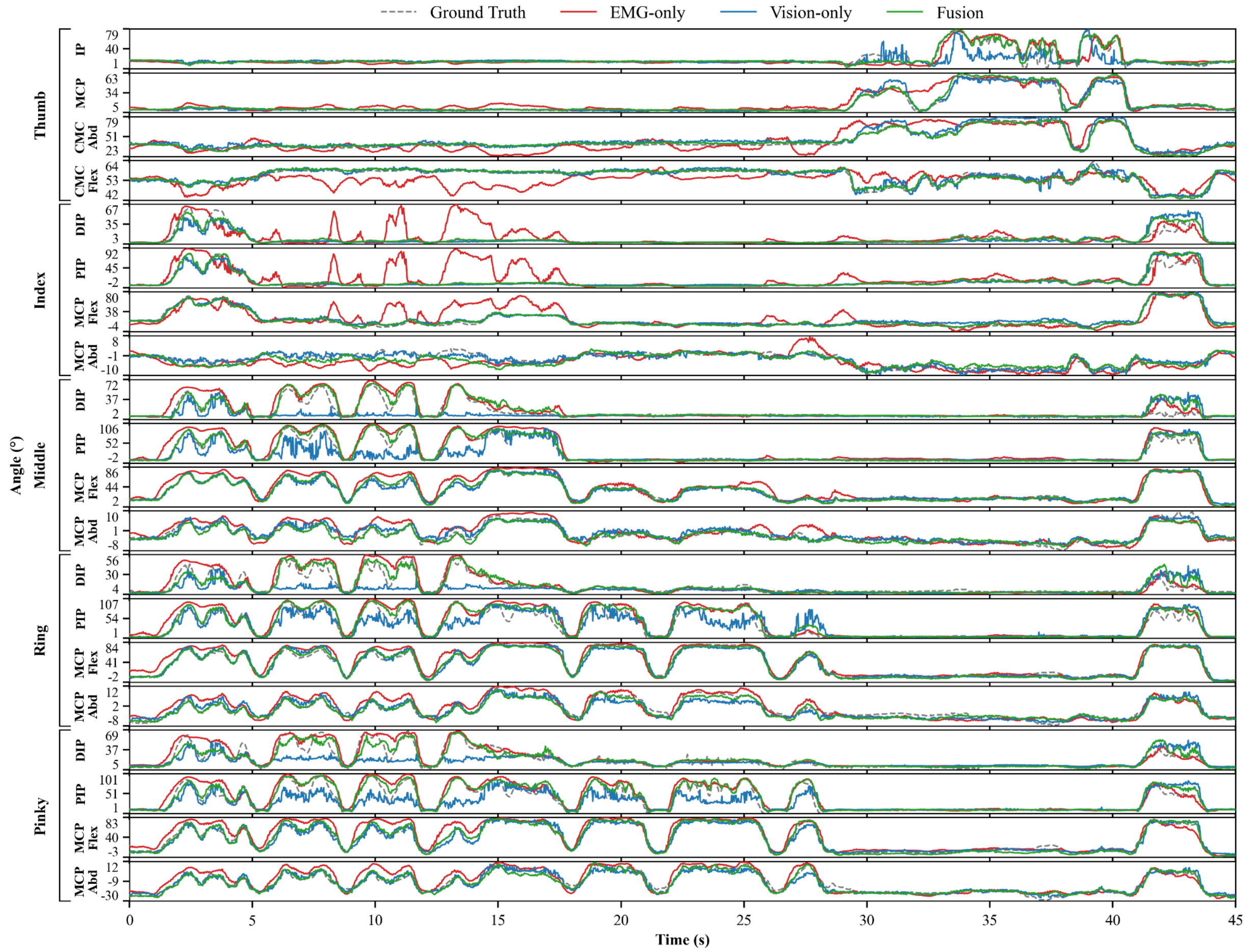

***Fig. 4***. *Temporal trajectories of predicted angles for 20 hand joint DoFs under different input modalities with the palm facing down. The fusion model follows the ground truth more closely and stably than the EMG-only and vision-only models.*

he EMG-only model captured the overall trends of finger motion but showed noticeable deviations in absolute joint angles. Its predictions for the index-finger DIP, PIP, and MCP-flexion DoFs exhibited additional peaks from approximately 7 to 18s that were absent from the ground-truth trajectories. Similar coupled fluctuations were also observed in several thumb-related DoFs. These patterns suggest that under periods of coordinated multi-finger movement, the EMG-only model may have difficulty distinguishing motions of different fingers and joint DoFs. Such ambiguity may reflect noise and cross-talk in surface EMG recordings, which can limit the separation of digit-specific motion information [22].

The vision-only model captured visually prominent changes in overall hand configuration, particularly for visible MCP flexion and abduction DoFs. However, errors were more apparent for distal joints, including the DIP and PIP joints, and for distal thumb DoFs such as the thumb IP joint. During some intervals, the vision-only predictions exhibited noise, temporal lag, or amplitude bias. This pattern is consistent with the greater susceptibility of distal phalanges to self-occlusion, finger overlap, and viewpoint limitations in egocentric images.

These qualitative observations are consistent with complementary information from the two modalities: vision provided spatial cues for visible joints, whereas EMG contributed motion-related information when visual estimation was less reliable. Accordingly, the fusion model followed the ground-truth trajectories more closely for distal joints, thumb-related DoFs, and complex gesture transitions.

As shown in Fig. 5, hand meshes were reconstructed from the predicted joint angles for six test snapshots using the mesh utilities in UmeTrack [23]. Each row shows the input egocentric RGB image and the corresponding meshes reconstructed from the ground-truth angle as well as fusion, vision-only, and EMG-only predictions.

Across these snapshots, the fusion model generally produced mesh configurations closer to the ground truth, particularly for poses involving substantial finger flexion or small separations between adjacent fingers. The vision-only model often captured the overall hand configuration and visible joints but showed local finger-pose errors when visual cues were less distinct, for example owing to occlusion, finger overlap, or motion blur. The EMG-only model captured coarse hand configurations but exhibited larger deviations in finger flexion and inter-finger configuration. These qualitative examples complement the quantitative modality-ablation results by illustrating the practical effect of multimodal fusion in visually challenging hand configurations.

Table 1 compares the regression performance of the proposed multimodal fusion model, the vision-only model, and the EMG-only model across hand DoFs under two palm orientations. Results are reported as mean $R^2$/MAE (°) across 10 participants. Overall, the fusion model achieved the best overall mean performance under both palm orientations. In the palm-facing-down condition, the fusion model achieved 0.88/2.82°, compared with 0.83/3.19° for the vision-only model and 0.68/5.03° for the EMG-only model. In the palm-facing-side condition, the corresponding values were 0.88/2.78°, 0.82/3.24°, and 0.68/5.24°, respectively.

For the thumb DoFs, performance varied across joints and palm orientations. In the palm-facing-down condition, the thumb IP joint showed lower vision-only performance ($R^2$/MAE = 0.55/3.59°) than fusion performance (0.70/3.33°). The EMG-only model also showed lower performance on the thumb IP joint (0.36/5.60°). Thumb IP flexion is primarily driven by a deep forearm flexor pollicis longus [24]. Its contribution may be less selectively represented in surface EMG recordings when activity from adjacent forearm muscles overlaps. Together with noise and cross-talk in surface EMG recordings [22], this may reduce the separability of thumb IP motion during coordinated hand movements. In contrast, fusion improved thumb IP estimation relative to both single-modality models. For the thumb MCP joint in the palm-facing-down condition, the fusion and vision-only models showed similar performance (0.97/1.40° and 0.97/1.47°, respectively). In the palm-facing-side condition, fusion was slightly lower than vision-only for the thumb IP joint (0.80/3.32° versus 0.82/3.05°) and comparable for the thumb MCP joint (0.90/1.93° versus 0.90/1.90°). These results indicate that the contribution of the two modalities varied across thumb joints and viewing conditions, with fusion providing its clearest benefit when single-modality estimates were less reliable. For the thumb CMC-related DoFs, including CMC_Abd and CMC_Flex, the fusion and vision-only models showed closely matched performance under both palm orientations. Fusion yielded small improvements for CMC_Flex, whereas CMC_Abd performance was nearly identical between the two models. The high vision-only performance for these DoFs indicates that visual information alone provided strong predictive cues in these conditions. EMG information nevertheless provided a small additional benefit for CMC_Flex.

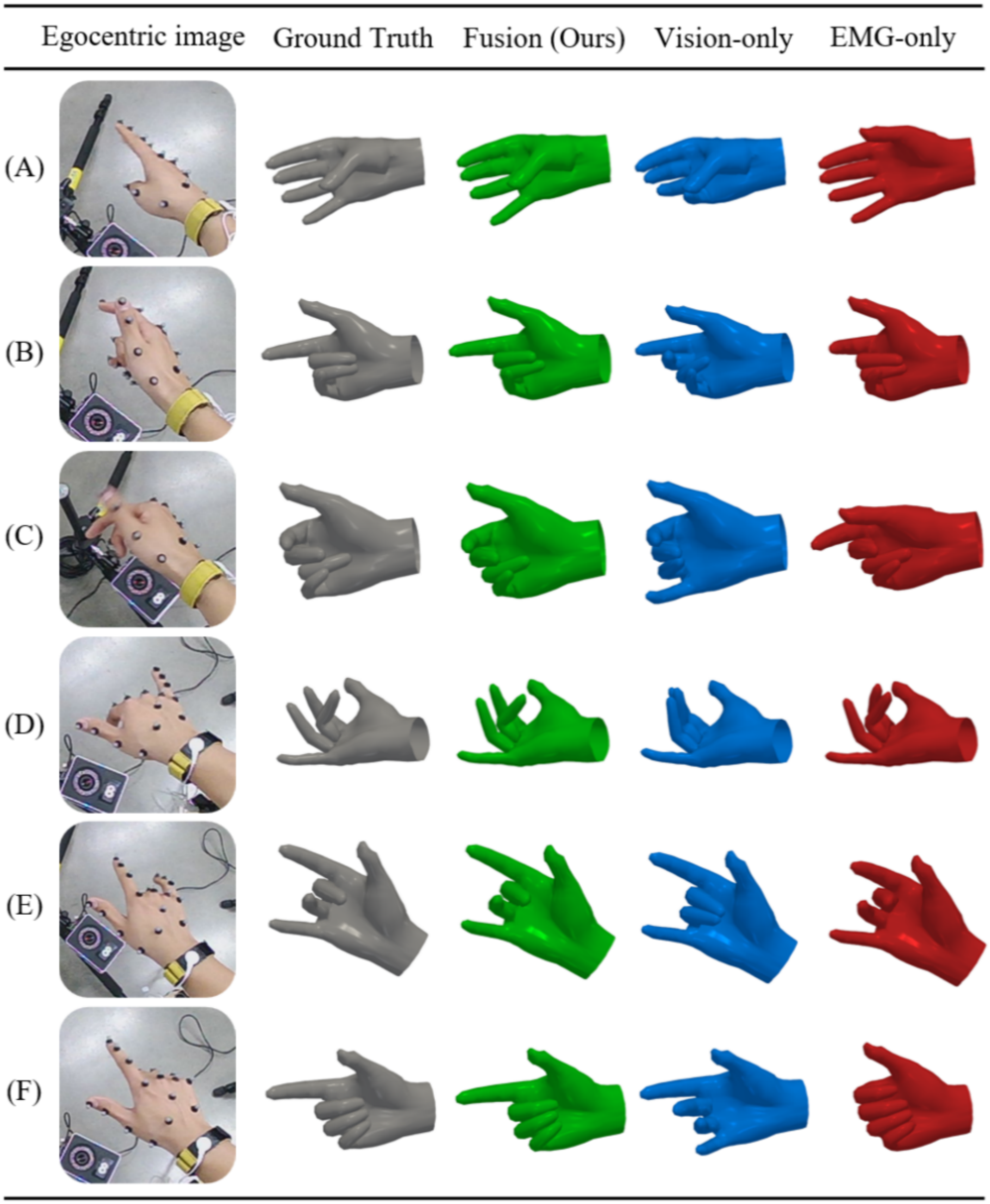

***Fig. 5**. Qualitative comparison of reconstructed hand meshes under different input modalities. From left to right, the columns show the egocentric image, ground-truth hand pose, fusion result, vision-only result, and EMG-only result. The fusion model produces hand poses that are closer to the ground truth.*

| Evaluation metrics: $R^2$ / MAE (°) | | Palm facing down | | | Palm facing side | | |
|---|---|---|---|---|---|---|---|
| Finger | Joint | Fusion | Vision-only | EMG-only | Fusion | Vision-only | EMG-only |
| Thumb | IP | **0.70 / 3.33** | 0.55 / 3.59 | 0.36 / 5.60 | 0.80 / 3.32 | **0.82 / 3.05** | 0.44 / 5.84 |
| | MCP | **0.97 / 1.40** | 0.97 / 1.47 | 0.73 / 4.24 | 0.90 / 1.93 | **0.90 / 1.90** | 0.57 / 3.89 |
| | CMC_Abd | **0.99 / 1.22** | 0.98 / 1.24 | 0.71 / 5.32 | 0.99 / 1.52 | **0.99 / 1.51** | 0.77 / 5.25 |
| | CMC_Flex | **0.96 / 0.66** | 0.96 / 0.71 | 0.45 / 2.47 | **0.95 / 0.92** | 0.95 / 0.96 | 0.56 / 2.67 |
| Index | DIP | **0.87 / 2.45** | 0.85 / 2.65 | 0.62 / 4.94 | **0.97 / 1.69** | 0.96 / 1.76 | 0.63 / 5.82 |
| | PIP | **0.93 / 2.77** | 0.92 / 3.01 | 0.74 / 6.99 | **0.98 / 1.76** | 0.98 / 1.94 | 0.69 / 7.72 |
| | MCP_Flex | **0.98 / 1.74** | 0.98 / 1.80 | 0.74 / 7.12 | 0.97 / 2.17 | **0.98 / 1.98** | 0.75 / 6.77 |
| | MCP_Abd | 0.75 / 1.14 | **0.81 / 0.99** | 0.24 / 2.01 | 0.71 / 1.86 | **0.72 / 1.76** | 0.26 / 3.07 |
| Middle | DIP | **0.73 / 3.49** | 0.58 / 4.34 | 0.58 / 4.70 | **0.95 / 2.49** | 0.92 / 3.04 | 0.79 / 4.78 |
| | PIP | **0.89 / 5.38** | 0.88 / 5.88 | 0.79 / 8.06 | **0.97 / 3.29** | 0.97 / 3.71 | 0.84 / 7.69 |
| | MCP_Flex | 0.98 / 2.27 | **0.98 / 2.26** | 0.91 / 4.91 | **0.95** / 2.80 | 0.94 / **2.78** | 0.85 / 4.81 |
| | MCP_Abd | 0.91 / 1.04 | **0.92 / 0.96** | 0.73 / 1.72 | **0.55 / 1.59** | 0.42 / 1.66 | 0.35 / 2.01 |
| Ring | DIP | **0.72 / 2.84** | 0.50 / 3.55 | 0.57 / 3.66 | **0.91 / 2.83** | 0.85 / 3.65 | 0.77 / 4.85 |
| | PIP | **0.86 / 7.12** | 0.81 / 8.55 | 0.80 / 8.97 | **0.97 / 4.43** | 0.94 / 5.30 | 0.84 / 8.69 |
| | MCP_Flex | **0.98 / 2.62** | 0.98 / 2.65 | 0.92 / 4.81 | **0.94 / 3.29** | 0.90 / 4.05 | 0.88 / 4.71 |
| | MCP_Abd | 0.92 / 1.06 | **0.92 / 1.02** | 0.77 / 1.76 | **0.78 / 1.71** | 0.55 / 2.31 | 0.73 / 2.02 |
| Pinky | DIP | **0.70 / 3.56** | 0.50 / 4.63 | 0.49 / 4.90 | **0.90 / 3.67** | 0.87 / 4.46 | 0.76 / 5.66 |
| | PIP | **0.83 / 6.45** | 0.71 / 8.41 | 0.71 / 9.04 | **0.93 / 5.45** | 0.87 / 6.95 | 0.81 / 8.46 |
| | MCP_Flex | **0.94 / 3.73** | 0.94 / 3.82 | 0.86 / 6.17 | **0.85 / 4.97** | 0.70 / 6.74 | 0.81 / 5.64 |
| | MCP_Abd | **0.91 / 2.13** | 0.90 / 2.18 | 0.80 / 3.31 | **0.60 / 3.90** | 0.23 / 5.29 | 0.53 / 4.46 |
| Overall Mean | | **0.88 / 2.82** | 0.83 / 3.19 | 0.68 / 5.03 | **0.88 / 2.78** | 0.82 / 3.24 | 0.68 / 5.24 |

***Table 1****. Comparison of average regression performance for Fusion, Vision-only, and EMG-only models across hand DoFs under different palm orientations, values are reported as $R^2$ / MAE (°). Bold values indicate the best value for each metric, i.e., the highest $R^2$ or the lowest MAE, for each DoF and palm orientation.*

For the index, middle, ring, and pinky fingers, fusion and vision-only performance was similar for several MCP-flexion DoFs in the palm-facing-down condition, whereas larger fusion gains were observed for several PIP and DIP DoFs. For example, for the middle finger, PIP performance improved from 0.88/5.88° with the vision-only model to 0.89/5.38° with fusion. The improvement for the DIP joint was larger, from 0.58/4.34° to 0.73/3.49°. These results are consistent with more challenging visual estimation for some distal joints in the palm-facing-down condition. In the palm-facing-side condition, vision-only performance was generally lower for ring and pinky flexion-related DoFs than for the index finger. For example, vision-only MCP-flexion performance decreased from 0.98/1.98° for the index finger to 0.70/6.74° for the pinky finger. Fusion improved pinky MCP-flexion performance to 0.85/4.97°, reducing this performance gap. The MCP_Abd DoFs also showed fusion gains in the palm-facing-side condition, particularly for the middle, ring, and pinky fingers. Middle-finger MCP_Abd performance improved from 0.42/1.66° with vision-only to 0.55/1.59° with fusion. Ring-finger MCP_Abd improved from 0.55/2.31° to 0.78/1.71°, and pinky-finger MCP_Abd improved from 0.23/5.29° to 0.60/3.90°. In contrast, vision-only was slightly better than fusion for index-finger MCP_Abd (0.72/1.76° versus 0.71/1.86°). In the side view, the middle, ring, and pinky fingers may be more susceptible to finger overlap and occlusion, while monocular RGB images do not provide explicit depth information. These factors can make abduction/adduction movements, which often involve depth-wise displacement, more difficult to estimate. EMG may provide complementary information for these visually constrained DoFs, contributing to the improved fusion performance observed for the middle, ring, and pinky MCP_Abd DoFs. Overall, the fusion model outperformed the single-modality models across most DoFs and both palm orientations. These results show that combining visual and EMG information improved hand joint-angle regression for visually challenging DoFs in this evaluation.

### 3.2 Comparison with State-of-the-Art Unimodal Methods

To further validate the effectiveness of the proposed fusion model, we compared it with representative state-of-the-art unimodal methods using the same data splits and evaluation metrics. The comparison methods include EMG-based models (emg2pose, NeuroPose) and vision-based models (HaMeR, WiLoR). To enable a common evaluation, the outputs of all methods were converted or adapted to the 20-DoF joint-angle representation used in this study. The experimental results are presented in Table 2, which reports the mean and standard deviation of $R^2$ and MAE across different participants.

Because the 128-channel HD-sEMG input, temporal window length, and output temporal resolution used in this study differ from the original configurations of emg2pose and NeuroPose, we adapted the two EMG-based baselines while retaining their principal network designs. For emg2pose, we adjusted the convolutional kernel sizes and strides in the TDS backbone to reduce excessive temporal compression and to match the temporal resolution required for the joint-angle regression task. For NeuroPose, we retained the encoder, residual blocks, and decoder structure, while adjusting the pooling and upsampling scales according to the temporal length and electrode-channel arrangement of the HD-sEMG input. For the vision-based baselines, we evaluated HaMeR and WiLoR, two open-source monocular hand-mesh reconstruction methods. Because the joint-angle labels in this study were derived from dorsal hand-marker trajectories recorded using an optical motion-capture system, their definitions do not exactly correspond to the MANO parameter space. We therefore evaluated two variants for each method. In the first variant, a MANO-to-angle MLP was trained to map the MANO parameters output by the vision model, together with the corresponding OpenPose-style hand-skeleton representation, to the 20-DoF joint-angle space used in this study. In the second variant, denoted HaMeR Backbone or WiLoR Backbone, the pretrained visual backbone was connected to the angle decoder used in this study and trained to directly regress the 20 joint angles from image features. This design evaluates the transferability of the visual representations while avoiding the additional mismatch introduced by mapping MANO-based outputs to the study-specific joint-angle definition.

| | Method | $R^2$ | MAE (°) |
|---|---|---|---|
| EMG Methods | NeuroPose [12] | 0.44 ± 0.12 | 7.73 ± 1.12 |
| | emg2pose [13] | 0.66 ± 0.08 | 5.38 ± 0.76 |
| | **EMG-only** | 0.68 ± 0.08 | 5.13 ± 0.77 |
| Vision Methods | HaMeR [9] | 0.75 ± 0.06 | 4.30 ± 0.53 |
| | WiLoR [11] | 0.76 ± 0.07 | 4.21 ± 0.56 |
| | **HaMeR Backbone (Vision-only)** | 0.82 ± 0.05 | 3.21 ± 0.36 |
| | WiLoR Backbone | 0.79 ± 0.06 | 3.61 ± 0.41 |
| | **Fusion (Ours)** | 0.88 ± 0.03 | 2.80 ± 0.30 |

***Table 2****. Comparison with state-of-the-art EMG and RGB vision-based unimodal methods. Bold entries indicate the EMG-only and vision-only models used for the modality ablation study in Section 3.1.*

As shown in Table 2, the proposed multimodal fusion model achieved the highest mean $R^2$ and the lowest mean MAE among all evaluated methods. Among the evaluated unimodal methods, HaMeR Backbone achieved the best performance among the vision-only methods, whereas our EMG-only variant performed best among the EMG-based methods. For both HaMeR and WiLoR, coupling the visual backbone with the angle decoder used in this study to directly regress joint angles outperformed mapping the original model outputs to joint angles using MANO-to-angle. This result suggests that direct regression from visual features may better align the learned representation with the study-specific joint-angle label definition. Although WiLoR slightly outperformed HaMeR when their original MANO outputs were mapped to joint angles, HaMeR Backbone outperformed WiLoR Backbone when both visual backbones were coupled to the angle decoder in this study. One possible factor is the representation design of WiLoR, which jointly processes image tokens with learnable pose, shape, and camera tokens to predict MANO and camera parameters. The resulting features may therefore retain task-specific structure associated with MANO recovery, making their transfer to the study-specific 20-DoF joint-angle regression task less direct. Based on the comparison above, we selected HaMeR as the visual backbone of the proposed fusion model. Overall, the comparative experiments demonstrate that the proposed multimodal fusion model can effectively integrate EMG signals and visual information, achieving better prediction accuracy and stability compared with state-of-the-art unimodal methods.

### 3.3 Comparison of EMG Encoder and Fusion Decoder Designs

To assess the effect of EMG encoder architecture on model performance, we compared three EMG encoders: MLP, bidirectional long short-term memory (BiLSTM), and Transformer. For a controlled comparison, all encoders used the same input features and were configured with the same hidden representation dimension and number of layers; only the main EMG encoder architecture was changed. The comparison was conducted in both the EMG-only and multimodal fusion settings to examine the effect of encoder selection on unimodal modeling and multimodal prediction. As shown in the upper section of Table 3, the Transformer-based model achieved the best performance in both the EMG-only and multimodal fusion settings. This performance advantage is consistent with the ability of self-attention to model temporal dependencies and relationships among EMG features and channels.

| **Encoder Method** | | $R^2$ | **MAE (°)** |
|---|---|---|---|
| | MLP | $0.63 \pm 0.09$ | $5.69 \pm 0.91$ |
| **EMG-only** | BiLSTM | $0.64 \pm 0.09$ | $5.51 \pm 0.80$ |
| | **Transformer (Ours)** | $0.68 \pm 0.08$ | $5.13 \pm 0.77$ |
| | MLP | $0.85 \pm 0.05$ | $3.04 \pm 0.34$ |
| **Fusion** | BiLSTM | $0.86 \pm 0.03$ | $3.06 \pm 0.33$ |
| | **Transformer (Ours)** | $0.88 \pm 0.03$ | $2.80 \pm 0.30$ |
| **Fusion Decoder Method** | | $R^2$ | **MAE (°)** |
| Late Fusion | | $0.87 \pm 0.03$ | $2.96 \pm 0.33$ |
| Bidirectional Cross-attention | | $0.86 \pm 0.04$ | $3.12 \pm 0.53$ |
| Global-query Transformer | | $0.86 \pm 0.04$ | $3.14 \pm 0.38$ |
| **DoF-query Transformer (Ours)** | | $0.88 \pm 0.03$ | $2.80 \pm 0.30$ |

***Table 3**. Ablation study of EMG encoder architectures and fusion decoder designs. The upper section compares MLP, BiLSTM, and Transformer encoders in the EMG-only and multimodal fusion settings. The lower section compares fusion decoder methods under the multimodal setting. Results are reported as mean ± standard deviation across 10 participants for $R^2$ and MAE (°). Bold values indicate the best performance within each comparison.*

To assess the influence of fusion decoder design on multimodal prediction performance, we compared Late Fusion, Bidirectional Cross-Attention, a global-query Transformer, and the proposed DoF-query Transformer. All methods used the same EMG encoder and vision encoder, as well as the same input representations and training protocol. The decoder depth and hidden dimension were matched wherever the architectures permitted a direct comparison. Late Fusion first predicted joint angles independently from the EMG-only and vision-only branches, and then combined the two unimodal predictions through per-DoF residual fusion at the output level. The Bidirectional Cross-Attention method introduced feature-level interaction between the modalities.

EMG and visual tokens attended to one another, using tokens from the other modality as key-value pairs. The resulting features were pooled into a multimodal representation for final regression. The proposed DoF-query Transformer assigned an independent learnable query to each DoF. Through decoder self-attention, the queries modeled relationships among DoFs. Through cross-attention, they extracted DoF-relevant information from the concatenated EMG and visual tokens. A regression head then produced a separate joint-angle prediction for each DoF. As an ablation, the global-query Transformer used a single global learnable query to extract fused features from the multimodal memory and jointly regress all DoFs.

As shown in the lower section of Table 3, the proposed DoF-query Transformer achieved the highest mean $R^2$ and the lowest mean MAE among the evaluated fusion decoders. Late Fusion achieved competitive performance. However, because it combines unimodal predictions only at the output level, it does not explicitly model feature-level interactions between EMG and visual information. Although Bidirectional Cross-Attention and the global-query variant introduced feature-level fusion, their performance remained below that of the proposed model, possibly because they may not fully accommodate the distinct multimodal information requirements of individual DoFs. These results suggest that assigning separate learnable queries to individual DoFs enables more effective DoF-specific multimodal fusion for the present joint-angle regression task.

### 3.4 DoF-Specific and Context-Adaptive Attention Analysis

To examine the DoF-specific and context-adaptive modality-attention patterns of the proposed DoF-query fusion decoder, we analyzed the cross-attention mass associated with the DoF queries. Fig. 6(A) and (B) show the EMG and visual attention mass across DoFs, together with the percentage changes in MAE of the proposed DoF-query Transformer and the global-query variant with respect to the single-modality baselines. Because the decoder memory contains substantially more visual tokens than EMG tokens, the absolute attention mass is affected by the unequal numbers of tokens in the two modalities. We therefore focus on relative variations in attention mass across DoFs and experimental conditions, rather than interpreting its absolute magnitude as a direct measure of modality reliability or contribution.

The attention patterns varied across DoFs. In the global-query model, all DoFs shared a single global-query and therefore followed a common modality-attention pattern. Its MAE reductions were uneven across DoFs, suggesting that a global attention pattern may not adequately represent the distinct multimodal information requirements of individual joints. In contrast, the proposed DoF-query model assigned a separate query to each DoF and achieved MAE comparable to or lower than that of the stronger single-modality baseline for most DoFs. Larger MAE reductions were observed when the complementary modality provided additional information for a particular DoF. These patterns suggest that separate DoF queries enabled more flexible cross-modal selection for joint-angle regression.

This observation is consistent with the DoF-wise results in Table 1. Under the palm-facing-down condition, several distal joints showed larger MAE reductions relative to the vision-only model, and these reductions were accompanied by relatively higher EMG attention mass. This pattern is consistent with the greater visual ambiguity of some distal joints under self-occlusion. Under the palm-facing-side condition, several middle-, ring-, and pinky-finger DoFs showed larger MAE reductions relative to the corresponding vision-only estimates, together with relatively higher

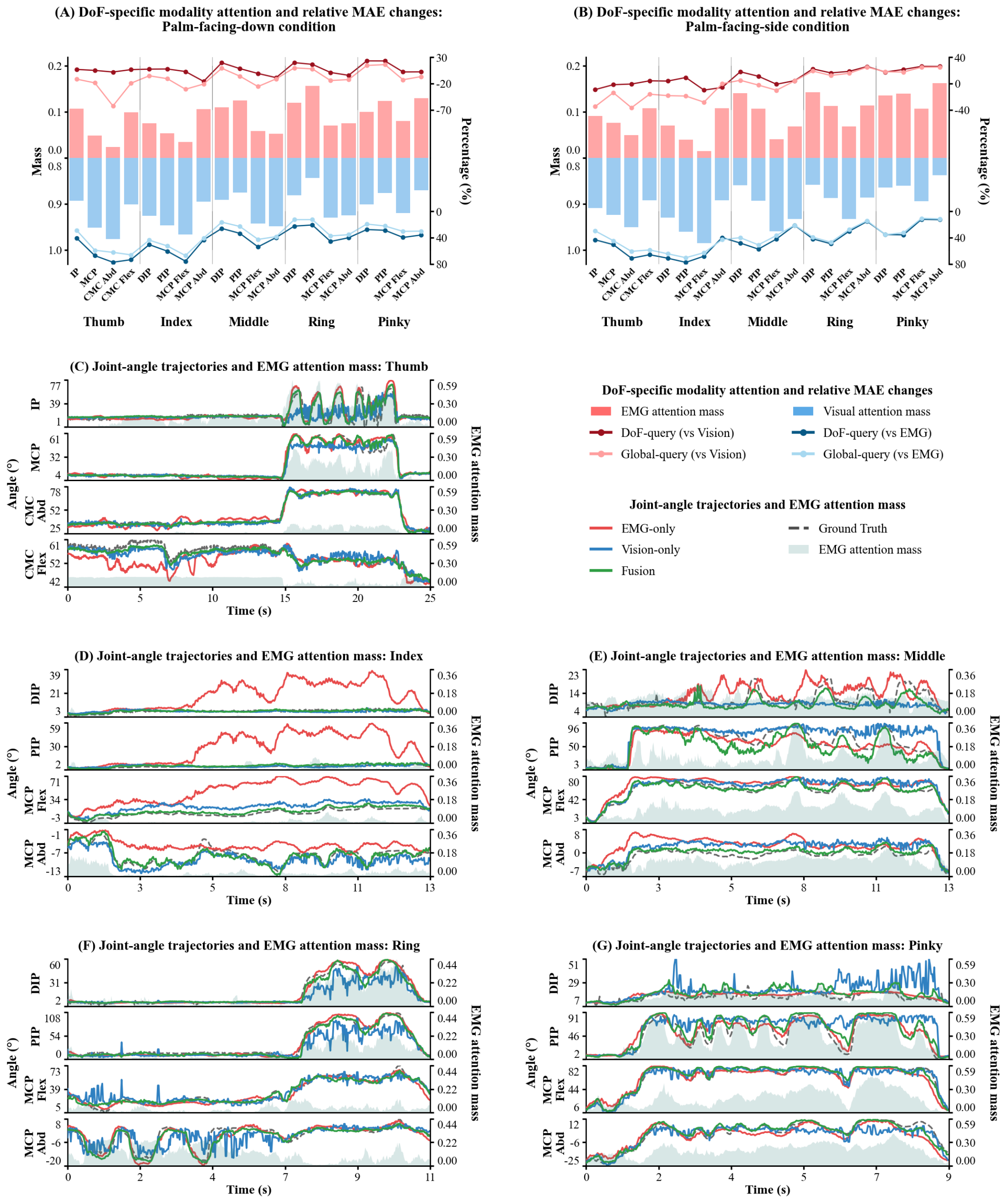

***Fig. 6**. DoF-specific modality attention patterns and representative temporal analyses of the proposed fusion model. (A, B) EMG and visual attention mass for individual hand DoFs, together with the relative MAE changes of the proposed DoF-query Transformer and the global-query variant with respect to the vision-only and EMG-only baselines, under the palm-facing-down and palm-facing-side conditions, respectively. Vision attention mass is displayed downward for visual comparison. (C)–(G) Ground-truth joint angles, EMG-only, vision-only, and fusion model predictions, and EMG attention mass for the thumb, index, middle, ring, and pinky finger, respectively, under representative motions. These examples illustrate that modality-attention patterns vary across DoFs and adapt dynamically to the temporal context.*

EMG attention mass in the fusion model. Finger overlap, occlusion, and depth-wise motion in the side view may reduce the available visual evidence for these DoFs, whereas EMG can provide complementary motion-related information.

Fig. 6(C) to Fig. 6(G) jointly illustrate representative predicted temporal trajectories and the corresponding changes in EMG attention mass across different hand-movement contexts. Here, context refers to the current combination of hand pose, motion phase, and visual observability reflected in the EMG and visual token representations. Fig. 6(C) shows a palm-facing-down movement. During the first 14 s, the participant performed isolated thumb flexion-extension, followed by repeated thumb-to-palm flexion, during which the thumb IP and MCP joints were partially obscured in the egocentric view. Fig. 6(F) shows a palm-facing-side movement in which all five fingers repeatedly transitioned between a fingers-together posture and a fully spread-finger posture during the first 6 s, followed by irregular, asynchronous flexion-extension movements of the four fingers. The side view provided limited depth cues for resolving abduction-related changes in relative finger position, while finger overlap and self-occlusion further reduced visual observability. In these visually challenging intervals, the vision-only model showed larger absolute errors and noticeable temporal jitter during several local transitions. In contrast, the EMG-only model captured the local movements more reliably, with smaller errors and closer temporal tracking of the ground-truth trajectories. The fusion model further reduced these errors and more closely tracked the ground-truth trajectories. EMG attention mass increased when vision-only estimation became less reliable and decreased as its reliability recovered.

Figs. 6(D) and 6(E) show a palm-facing-down movement in which the thumb and index finger remained relaxed, whereas the distal joints of the middle, ring, and little fingers underwent irregular but synchronized flexion-extension movements. In Fig. 6(D), the EMG-only predictions exhibited coupled deviations across joints, mainly as fluctuations in the nominally relaxed index finger. In comparison, the fusion model more clearly separated the stable and moving joint states. Meanwhile, the EMG attention mass for the nominally relaxed index finger remained close to zero. Together, these patterns are consistent with complementary modality use. When visual representations were less informative under the current visibility conditions or motion phase, the decoder placed greater attention on EMG features. Conversely, visual spatial cues helped constrain EMG-derived motion information and reduce cross-digit ambiguity during coordinated multi-finger motion. Thus, the two modalities provided distinct information that could compensate for limitations in the other modality.

During some transitions, such as those shown in Fig. 6(E), both unimodal models showed local deviations from the ground truth. Nevertheless, the fusion model remained closer to the ground-truth trajectory, suggesting that it could combine partially informative cues from both modalities rather than relying exclusively on the modality with lower standalone error. Fig. 6(G) shows a palm-facing-down movement involving repeated flexion and extension of the ring and little fingers. The EMG-only model reflected the timing and direction of several local changes but showed less accurate absolute angle estimates. In contrast, the vision-only model generally retained the approximate angle range and overall configuration, but its predictions became locally flat or showed noisy fluctuations during several transitions. Thus, neither modality alone fully described these local changes. By combining trend-related cues from EMG with the spatial constraints provided by vision, the fusion model generally followed the ground-truth trajectories more closely.

During intervals in which the vision-only trajectories became locally flat or deviated from the ground truth, the EMG attention mass increased. This pattern is consistent with the decoder placing greater emphasis on EMG features when they supplied complementary dynamic information in the current movement context.

Overall, the query-level analysis suggests that the proposed decoder learned DoF-specific and context-adaptive modality-attention patterns rather than relying on a single fixed global preference. Compared with the global-query variant, the separate DoF queries provided a more interpretable view of how the decoder selectively used EMG and visual features for different joint-angle targets under changing hand configurations, visibility conditions, and motion phases. These attention patterns should be interpreted as model-internal indicators of selective feature use, rather than as direct measurements of modality reliability or explicit detection of modality failure. Nevertheless, together with the ablation analyses, they support the effectiveness of DoF-specific and context-adaptive multimodal fusion in the present task.

### 3.5 Robustness Evaluation under Synthetic Visual Occlusion

To evaluate the robustness of the models under degraded visual input, we simulated external visual occlusion during the testing stage. All models were trained on the original clean training set without artificial occlusion, while different levels of occlusion were applied only to RGB images during testing, in order to assess the generalization ability of the models to unseen visual disturbances. Specifically, occlusion was generated within the hand bounding box in the original image, and the occlusion ratio was defined as the number of occluded pixels divided by the total number of pixels inside the bounding box. The occlusion ratio was increased from 0% to 70%. A mixed occlusion strategy was adopted, including random-angle line-based occlusion and randomly rotated rectangular occlusion, where the occluded pixels were replaced by a random solid color. The occlusion operation was performed before image preprocessing. We compared vision-only, EMG-only, and multiple vision-EMG fusion methods described in Section 3.3, including Late Fusion, Bidirectional Cross-Attention, and the proposed DoF-query Transformer used in this work.

The experimental results in Fig. 7 show that the vision-only model is highly sensitive to visual occlusion. As the occlusion ratio increases, its $R^2$ decreases rapidly and becomes close to or below zero after approximately 30% occlusion, while the MAE increases substantially. This indicates that when the hand region in the visual input is corrupted, a model relying solely on RGB images struggles to maintain reliable joint-angle estimation. In contrast, the proposed fusion model consistently outperforms the vision-only model across all occlusion ratios and exhibits a slower performance degradation trend. This suggests that EMG signals can provide effective complementary muscle-activation information under degraded visual conditions, thereby significantly improving occlusion robustness compared with the visual unimodal model.

Compared with the EMG-only baseline, the fusion model shows a clear advantage under low occlusion levels. At 0% occlusion, the DoF-query Transformer achieves the highest $R^2$ and the lowest MAE, indicating that when visual information is reliable, the complementary fusion of vision and EMG can outperform either single modality. Under mild occlusion, the fusion model can still maintain performance close to or better than EMG-only. However, as the occlusion ratio further increases, the performance of the fusion model gradually falls below the EMG-only baseline. This phenomenon suggests that under severe occlusion, because the model cannot completely ignore the unreliable visual modality, corrupted visual features may act as noise and negatively affect the fusion process.

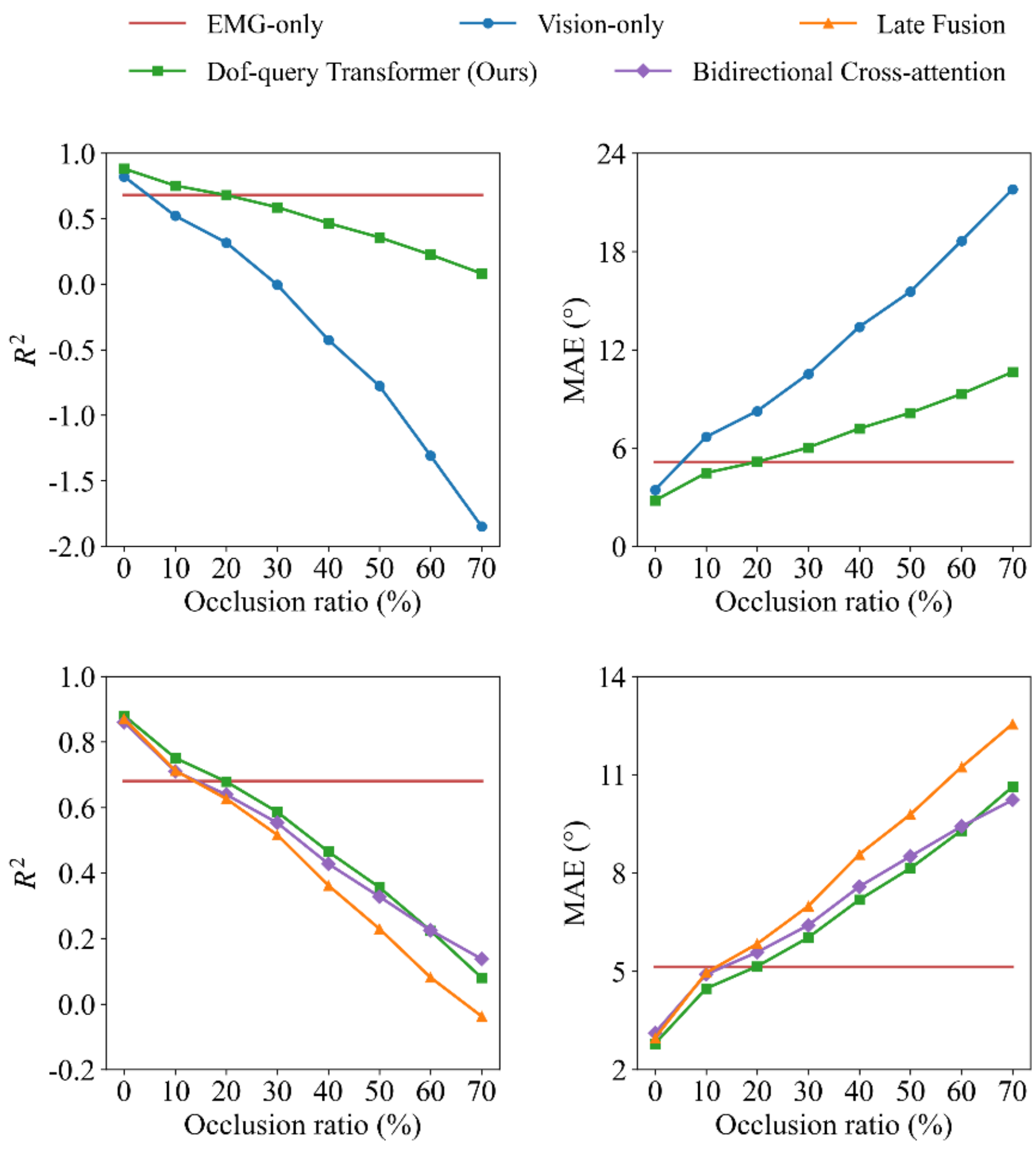

***Fig.** 7. Robustness evaluation under synthetic visual occlusion. Average performance of vision-only, EMG-only, and the fusion models is compared across increasing occlusion ratios from 0% to 70%, using $R^2$ and MAE as evaluation metrics.*

Among different fusion methods, the DoF-query Transformer demonstrates the most stable overall performance. It achieves the best performance under clean and low-occlusion conditions. Under moderate-to-high occlusion levels, compared with Late Fusion and Bidirectional Cross-Attention, its $R^2$ decreases more gradually and its MAE increases more slowly. Late Fusion shows the most obvious degradation as occlusion becomes stronger, indicating that simple late-stage fusion is insufficient to effectively handle corrupted visual features. Bidirectional Cross-Attention shows a slight advantage under high occlusion, but its overall trend is still weaker than that of the proposed method under low-to-moderate occlusion levels. Overall, the DoF-query Transformer can more effectively exploit the complementary relationship between EMG and visual information.

## 4. CONCLUSION

This study presented a finger-unencumbered multimodal framework for continuous 20-DoF hand joint-angle estimation using egocentric RGB images and 128-channel forearm HD-sEMG signals. The framework combines a ViT-based visual encoder, a Transformer-based EMG encoder, and a DoF-query Transformer fusion decoder. By assigning a separate learnable query to each joint DoF, the decoder performs DoF-specific and context-adaptive fusion of visual and EMG representations for joint-angle prediction.

Experiments on a synchronized multimodal dataset showed that the proposed fusion model outperformed the vision-only and EMG-only variants across most DoFs and palm orientations, with particularly clear benefits for visually challenging DoFs. The model also achieved better overall performance than the evaluated representative unimodal methods. The encoder and decoder ablation studies further supported the selection of the Transformer EMG encoder and the proposed DoF-query fusion decoder. Moreover, the query-level analysis indicated that the decoder learned DoF-specific and time-varying modality dependencies, rather than applying a single fixed fusion pattern to all joints.

The synthetic visual-occlusion experiments further showed that the proposed model was more resilient to degraded visual input than the vision-only model and the alternative fusion strategies under clean-to-moderate occlusion levels. However, performance gradually fell below that of the EMG-only model under severe occlusion. This result highlights a current limitation: although the proposed decoder adaptively fuses the two modalities, it does not explicitly estimate modality reliability or detect modality failure during fusion. Future work will investigate explicit modality-quality assessment to suppress corrupted visual information and enable more robust operation under severe occlusion or other sensor degradations. More broadly, the proposed framework provides a potential sensing pathway for capturing fine-grained human hand kinematics without instrumenting the fingers. When synchronized with egocentric observations, accurate continuous hand-kinematic trajectories could serve as dense action annotations for human manipulation demonstrations, supporting the construction of embodied-intelligence datasets, and the learning of dexterous robotic manipulation from human behavior.

## SUPPLEMENTARY MATERIALS

### 1. Calculation of Hand Joint Angles from Motion-Capture Marker Coordinates

The experimental configuration, marker placement, and motion-capture acquisition protocol are described in the Experimental Setup section of the main text. This section presents the procedure used to convert the three-dimensional marker trajectories into hand joint angles. All vector operations were performed in a right-handed coordinate system, and the resulting joint angles were expressed in degrees. The normalization operator $\mathcal{N}(\cdot)$ is defined as:

$$\mathcal{N}(\mathrm{x}) = \mathrm{x}/\parallel \mathrm{x} \parallel \quad (17)$$

First, we calculate the bone vectors of fingers, the thumb bone vectors were calculated as follows:

$$V_{CMC,1} = \mathcal{N}\left(f_{MCP,1} - f_{CMC,1}\right) \quad (18)$$

$$V_{MCP,1} = \mathcal{N}(f_{IP,1} - f_{MCP,1}) \tag{19}$$

$$V_{IP,1} = \mathcal{N}(f_{TIP,1} - f_{IP,1}) \tag{20}$$

The bone vectors of the other 4 fingers were calculated as follows:

$$V_{MCP,i} = \mathcal{N}(f_{PIP,i} - f_{MCP,i}) \tag{21}$$

$$V_{PIP,i} = \mathcal{N}(f_{DIP,i} - f_{PIP,i}) \tag{22}$$

$$V_{DIP,i} = \mathcal{N}(f_{TIP,i} - f_{DIP,i}) \tag{23}$$

$$i = 2,3,4,5$$

where $f_{J,K}$ denotes the marker placed at anatomical landmark J of finger k (k = 1 for the thumb, k = 2,3,4,5 for the index, middle, ring, and pinky fingers). Each bone vector is the unit vector pointing from the proximal marker to the distal marker of the corresponding phalanx.

Then, we calculate the reference vectors that define the palm coordinate frame and serve as references for joint angles that are not simple hinge rotations:

$$d_{palm} = \mathcal{N}\left(p_1 - \frac{p_2 + p_3}{2}\right) \tag{24}$$

$$n_{palm} = \mathcal{N}((p_2 - p_1) \times (p_3 - p_2)) \tag{25}$$

$$r_{abd} = \mathcal{N}(n_{palm} \times d_{palm}) \tag{26}$$

$$V_{IM} = \mathcal{N}(f_{MCP,2} - f_{CMC,1}) \tag{27}$$

$$n_{thumb} = \mathcal{N}(V_{IM} \times V_{CMC,1}) \tag{28}$$

where $p_1, p_2, p_3$ are the three palm markers (Palm 1, 2, 3). $d_{palm}$ is the palm direction pointing toward the fingers; $n_{palm}$ is the palm normal pointing dorsally; $r_{abd}$ lies in the palm plane and is perpendicular to $d_{palm}$, serving as the reference direction for finger abduction/adduction; $V_{IM}$ is the index metacarpal vector pointing from the thumb CMC to the index MCP marker, used as the reference for thumb CMC flexion/extension; $n_{thumb}$ is the normal of the thumb metacarpal plane, used to quantify thumb CMC abduction/adduction.

The joint angles are computed using three operators:

$$\angle_{vv}(u, w) = atan2(\|u_\perp \times w_\perp\|, u_\perp \cdot w_\perp) \tag{29}$$

$$\angle_{vp}(v, n) = arcsin(\mathcal{N}(v) \cdot \mathcal{N}(n)) \tag{30}$$

$$\angle_{pp}(n_1, n_2) = \arccos(\mathcal{N}(n_1) \cdot \mathcal{N}(n_2)) \tag{31}$$

Where $\angle_{vv}$ is the angle between two adjacent bone vectors projected onto the plane perpendicular to the flexion axis ( $u_{\perp}, w_{\perp}$ are the projected vectors), evaluated with atan2 for numerical stability over the full angular range; $\angle_{vp}$ is the signed angle between a vector and the plane with normal n; $\angle_{pp}$ is the angle between two planes given by their normals $n_1$ and $n_2$.

The thumb hand joint angles were obtained as follows:

$$\theta_{IP,1} = \angle_{vv}(V_{MCP,1}, V_{IP,1}) \tag{32}$$

$$\theta_{MCP,1} = \angle_{vv}(V_{CMC,1}, V_{MCP,1}) \tag{33}$$

$$\theta_{CMC,1}^{flex} = \angle_{vv}(V_{IM}, V_{CMC,1}) \tag{34}$$

$$\theta_{CMC,1}^{abd} = \angle_{pp}(n_{palm}, n_{thumb}) \tag{35}$$

The hand joint angles of the other 4 fingers were obtained as follows:

$$\theta_{PIP,i} = \angle_{vv}(V_{MCP,i}, V_{PIP,i}) \tag{36}$$

$$\theta_{DIP,i} = \angle_{vv}(V_{PIP,i}, V_{DIP,i}) \tag{37}$$

$$\theta_{MCP,i}^{flex} = \angle_{vp}(V_{MCP,i}, -n_{palm}) \tag{38}$$

$$\theta_{MCP,i}^{abd} = \angle_{vp}(V_{MCP,i}, r_{abd}) \tag{39}$$

$$i = 2,3,4,5$$

Together, Eq. (17)-(39) produce the total 20 hand joint angles: four angles for the thumb, including the interphalangeal (IP) flexion angle, the metacarpophalangeal (MCP) flexion angle, the carpometacarpal (CMC) flexion angle (CMC_Flex), and the CMC abduction/adduction angle (CMC_Abd); and four angles for each of the index, middle, ring, and pinky fingers, including the distal interphalangeal (DIP) flexion angle, the proximal interphalangeal (PIP) flexion angle, the MCP flexion angle (MCP_Flex), and the MCP abduction/adduction angle (MCP_Abd).

## 2. Predefined Gesture Movements in Data Collection

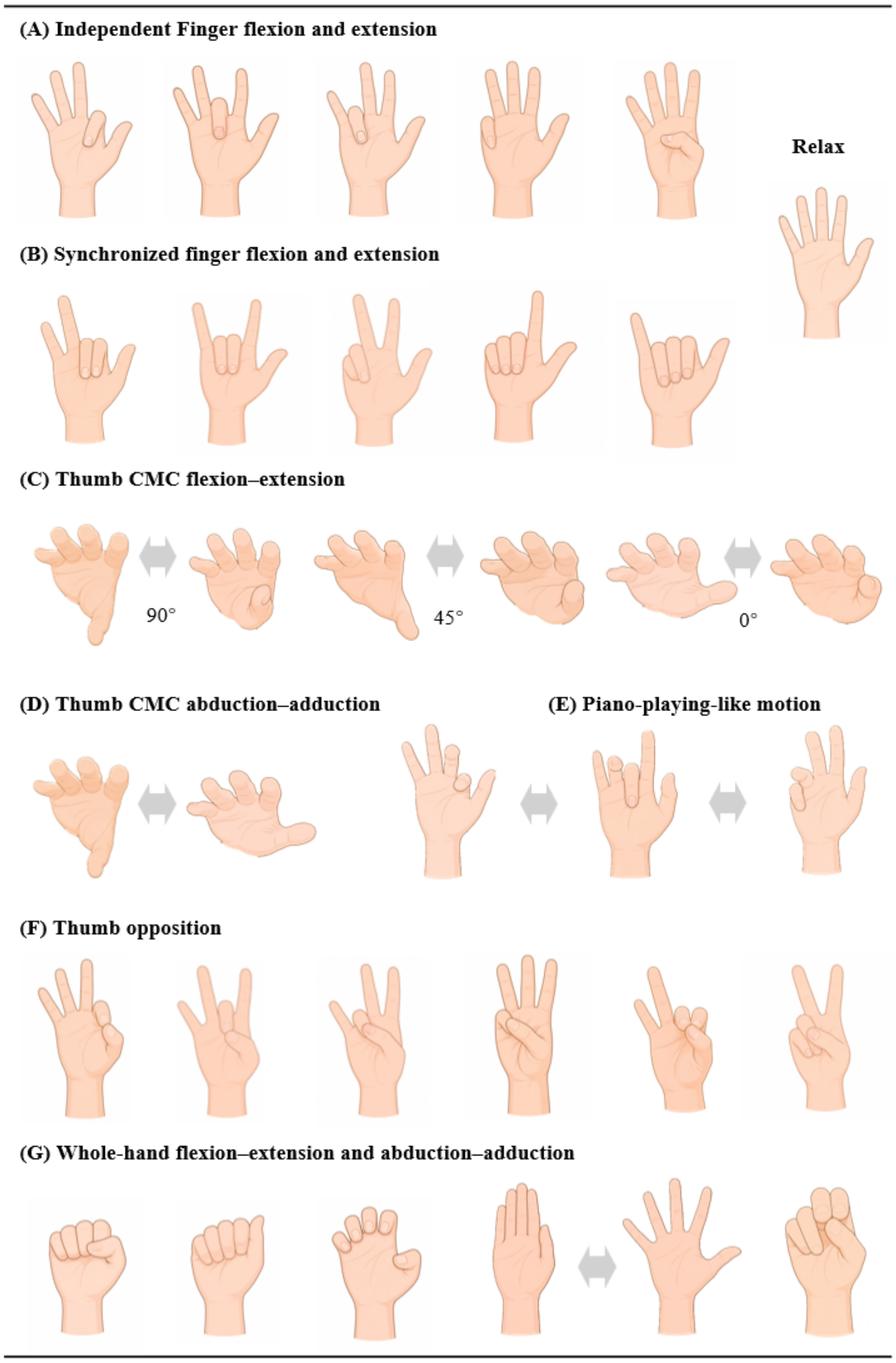

***Fig. S1***. *The 26 predefined gesture movements used for HD-sEMG data collection. The gesture set was organized into seven categories representing distinct hand movement synergy states: (A) independent finger flexion and extension, in which each finger flexes and extends in isolation; (B) synchronized finger flexion and extension, in which multiple fingers flex and extend together; (C) thumb carpometacarpal (CMC) flexion–extension performed at three abduction levels (90°, 45°, and 0°); (D) thumb CMC abduction–adduction; (E) piano-playing-like independent finger movements, in which several fingers move concurrently yet without coupling; (F) thumb opposition, in which the thumb sequentially opposes the other fingers; and (G) whole-hand flexion–extension and abduction–adduction. Bidirectional arrows indicate continuous transitions between paired postures.*